\documentclass[10pt,twocolumn,letterpaper]{article}

\usepackage[pagenumbers]{cvpr} % To force page numbers, e.g. for an arXiv version

\PassOptionsToPackage{table,dvipsnames}{xcolor}

\usepackage{graphicx}
\usepackage{subcaption} % or \usepackage{subfigure} if you must use the older package
\usepackage{hyperref}
\usepackage{url}
\usepackage{tabularx}
\usepackage{array}
\usepackage{colortbl}
\usepackage{calc}

\usepackage{atbegshi}% http://ctan.org/pkg/atbegshi
\definecolor{cvprblue}{rgb}{0.21,0.49,0.74}
\def\confName{CVPR}
\def\confYear{2025}

\title{Scalable Strategies for Continual Learning with Replay}

\author{Truman Hickok\\
Southwest Research Institute\\
San Antonio, Texas\\
{\tt\small truman.hickok@swri.org}
}

\begin{document}
\maketitle
\begin{abstract}
Future deep learning models will be distinguished by systems that perpetually learn through interaction, imagination, and cooperation, blurring the line between training and inference. This makes continual learning a critical challenge, as methods that efficiently maximize bidirectional transfer across learning trajectories will be essential. Replay is on track to play a foundational role in continual learning, allowing models to directly reconcile new information with past knowledge. In practice, however, replay is quite unscalable, doubling the cost of continual learning when applied naively. Moreover, the continual learning literature has not fully synchronized with the multi-task fine-tuning literature, having not fully integrated highly scalable techniques like model merging and low rank adaptation into a replay-enabled toolset that can produce a unified model in the face of many sequential tasks. In this paper, we begin by applying and analyzing low rank adaptation in a continual learning setting. Next, we introduce consolidation, a phasic approach to replay which leads to up to 55\% less replay samples being needed for a given performance target. Then, we propose sequential merging, an offshoot of task arithmetic which is tailored to the continual learning setting and is shown to work well in combination with replay. Finally, we demonstrate that the developed strategies can operate synergistically, resulting in a highly scalable toolset that outperforms standalone variants.
\end{abstract}    

\section{Introduction}
\label{sec:intro}

The trajectory of artificial intelligence is increasingly pointing towards systems capable of perpetual interaction and imagination within dynamic environments. Such systems are envisioned to continuously refine their understanding of the world, thereby generating progressively higher quality datasets tailored to their evolving capabilities \cite{pmlr-v235-hughes24a, silver2025experience, liang2024eurekaverse}. This paradigm shift necessitates not only the scaling of effective compute under existing frameworks, such as those for reinforcement learning \cite{guo2025deepseek, snell2024scaling} and pre-training \cite{wang2025scaling, dubey2024llama3}, but also an expansion of effective compute along the temporal dimension. Addressing this temporal aspect of learning is paramount for creating truly adaptive and self-improving AI.

Many contemporary approaches attempt to grapple with the temporal dimension by leveraging techniques like retrieval augmentation \cite{papagiannis2024rx, huang2024survey} and/or in-context learning \cite{dong2022survey, moeini2025survey}. A prominent example of this is the utilization of "skill libraries," often defined within a common programming language, which allow models to access and apply previously learned procedures at inference time \cite{wang2023voyager, wan2023lotus}. However, the field of continual learning (CL) offers a fundamentally distinct perspective by enabling models to dynamically and persistently modify their internal weights in response to new information \cite{wang2023comprehensive, verwimp2024continual}. These algorithms present an opportunity to amortize the learning cost—paying an upfront computational investment during training rather than continually expending resources on test-time adaptation. More profoundly, this weight modification allows new knowledge to be deeply integrated, reorganizing and extending existing manifolds. Such integration can lead to enhanced performance on novel tasks and foster a synergistic relationship where new information enriches, and is enriched by, existing and future knowledge \cite{yoon2023continual}.

Successfully implementing continual learning in this manner, especially for open-ended and self-improving systems, demands highly efficient methodologies. The challenge intensifies as models and datasets grow in scale, making it imperative to develop approaches that become more, rather than less, effective under such conditions. Replay, a straightforward yet powerful strategy, has been widely used \cite{wang2023comprehensive, ibrahim2024simple, garg2023ticclip}; nevertheless, the research community has largely overlooked strategies to minimize replay usage per unit of performance, an optimization path critical for scalable continual learning. Concurrently, two other families of algorithms, namely low-rank adaptation (LoRA) \cite{hu2021lora} and model merging \cite{ilharco2022editing, yang2024model}, have demonstrated significant promise as efficient and effective regularization techniques within the fine-tuning and multi-task learning literature. Despite their potential, their application and analysis within the continual learning setting remain largely unexplored, with LoRA, for instance, having only been cursorily studied in task-incremental learning (TIL) \cite{wistuba2023continual, wang2024hidepet, zhang2025clora, Yu_2024_CVPR}, a setting with distinct assumptions from the broader continual learning challenge \cite{vandeven2019three}.

In this paper, we study three families of continual learning algorithms that we identify as possessing high potential for scalability: low-rank adaptation, a novel “consolidation” strategy, and model merging. First, we conduct a comparative analysis of low-rank adaptation against full fine-tuning across diverse continual learning scenarios and varied data distributions. Our findings reveal that low-rank adaptation can drastically improve performance in highly under-regularized regimes (e.g when the data stream is hyper-compartmentalized or when the amount of replay samples per training batch is reduced) but is otherwise outperformed by full fine-tuning thanks to the effectiveness of replay and self-distillation. Second, we show that the sample efficiency of replay can be dramatically improved by reducing the number of replay samples in each training batch (optionally using low rank adaptation) then allocating training steps to a post-task "consolidation" period where the model is trained on a targeted distribution of samples; this leads to a 55\% reduction in the total number of replay samples required for a given performance target. Third, we adapt and apply model merging techniques to the continual learning context. We show that merging the model's parameters before and after learning a new task is particularly well-suited for the sequential bouts of learning seen in our problem setting, achieving performance comparable to maintaining an exponential moving average of the model's parameters throughout the entire training process but with greater efficiency and flexibility. Finally, we show that each strategy can be synergistically combined, ultimately resulting in a composite algorithm that outperforms each standalone version and reaches the same performance as our baseline with up to 65\% less replay samples.

\section{Related Work}
\label{sec:related}

\noindent\textbf{Continual Learning.} Continual learning (CL) methods have historically fallen into three broad categories \cite{hadsell2020embracing, wang2023comprehensive}: regularization-based, model expansion, and replay-based approaches. Regularization methods mitigate forgetting by constraining weight updates to preserve previously learned knowledge \cite{Kirkpatrick2017overcoming, smith2022closer, ke2023continual, jiang2025unlocking}. Model expansion techniques, by contrast, add new parameters or modules for each task to avoid interference (e.g., progressive networks and dynamic architectures) \cite{rusu2016progressive}. These can prevent forgetting entirely but may grow the model size unbounded. Replay-based (rehearsal) methods explicitly maintain a subset of past data (or a generative model thereof) and intermix those samples with current task data during training \cite{wang2023comprehensive}. In this paper, we assume that replay will remain a core component of CL algorithms. 

The concept of experience replay in CL is loosely inspired by neuroscience \cite{hayes2021replay, kumaran2016what}. Complex organisms are known to rehearse past and future experiences during offline periods (e.g sleep, daydreaming) in a highly strategic manner. These insights have begun to inform continual learning methods: the recent SIESTA algorithm \cite{harun2023siesta} introduces a dedicated “sleep” phase to consolidate knowledge between online learning sessions. SIESTA’s wake-sleep framework resembles our approach in spirit, though a key difference is that SIESTA entirely avoids replay during its online (wake) phase, whereas we employ a non-zero replay ratio and apply LoRA circumstantially while learning new tasks.

\noindent\textbf{Parameter-Efficient Fine-Tuning.} Another line of research relevant to CL is parameter-efficient fine-tuning (PEFT), which introduces small, task-specific parameter subsets to adapt a model instead of modifying all weights \cite{han2024parameter}. Techniques such as adapters \cite{houlsby2019parameter}, low-rank adaptation (LoRA) \cite{hu2021lora}, and prompt tuning \cite{lester-etal-2021-power} fall into this category. Originally popularized in natural language processing for multi-task and transfer learning, these methods have recently been applied to continual learning. Prompt-based approaches like L2P \cite{wang2022l2p} and DualPrompt \cite{wang2022dualprompt} freeze the backbone network and learn a set of prompts for each task, achieving competitive performance without storing replay data. Likewise, LoRA injects learned low-rank weight updates at each layer, greatly reducing the number of parameters that change per task. Early explorations of LoRA in vision CL (e.g., the CoLoR \cite{wistuba2023continual} method) report that LoRA-based adapters can outperform prompt-based tuning on class-incremental benchmarks. Importantly, these approaches operate in a task-incremental setting: each task gets its own adapter or prompt set, and a task ID is required to select the correct parameters at inference \cite{wistuba2023continual, wang2024hidepet, zhang2025clora, Yu_2024_CVPR}. The adapters themselves remain isolated per task and are not merged into the base model. This modular strategy can prevent interference, but it limits flexibility in class-incremental scenarios where task identity is unknown and all knowledge must reside in a single model. The present work builds on PEFT methods by examining how LoRA adapters can be integrated into a unified continual learner. By periodically merging the accumulated low-rank updates, we aim to retain the regularization and efficiency of PEFT while enabling seamless transfer and inference across tasks without oracle task information.

\noindent\textbf{Model Merging.} Model merging has emerged as a promising technique for multi-task fine-tuning in the era of large pre-trained models \cite{yang2024model, Yadav2024WhatMF}. Sometimes referred to as “task arithmetic,” model merging is a post-hoc strategy that composes multiple task-specialized models into one set of weights. Instead of co-training tasks (which risks interference), one can train separate models on each task and then combine their parameters. Prior work has shown that surprisingly simple merging schemes — such as weight averaging or selecting the largest-magnitude weight changes from each task — can recover most of the performance of individual models on their respective tasks. Moreover, merging typically leaves the original pre-trained knowledge intact, especially when using frontier models \cite{Yadav2024WhatMF} and algorithms like DARE-TIES \cite{Yu2024LanguageMA, Yadav2023TIESMergingRI} or Model Tailor \cite{zhu2024modeltailor}. These properties make model merging an attractive fit for continual learning, where we face a sequence of many tasks and wish to preserve all learned skills. 

In general, the efficacy of model merging tends to increase with model size and diversity of the pre-training corpus, suggesting that as we scale up continual learning (in terms of network capacity and number of tasks), merging strategies could become even more powerful \cite{Yadav2024WhatMF}. Our work contributes to this line of inquiry by adapting model merging to the continual learning setting: after each task’s training, we merge the model’s pre- and post-task weights (a form of sequential merge). In combination with replay and PEFT-based techniques, such merging offers a highly scalable way to combat forgetting in continual learning.

\section{Background}
\label{sec:background}

In this section, we outline our basic experimental setup, which is common to all upcoming experiments.

First, we pre-train a vision transformer (from the OpenCLIP library \cite{ilharco_gabriel_2021_5143773}) on a random 600-class subset of ImageNet-1k \cite{russakovsky2015imagenet}. Then, we fine-tune the model on a sequence of $T$ tasks, each with $C$ classes.

We study three problem settings that are common in continual learning: task-incremental learning (TIL), class-incremental learning (CIL), and continual pre-training (CPT) \cite{vandeven2019three, ke2023continual}. For TIL, we train a separate model for each of the $T$ tasks then report average accuracy across tasks ("1:T"). For CIL, we fine-tune the pre-trained model sequentially on $T$ tasks, reporting average accuracy across tasks ("1:T") after the entire sequence has been learned. For CPT, we seek to maximize performance on the pre-training dataset along with performance on the $T$ tasks; here, we report pre-training accuracy ("PT") and average accuracy across the $T$ tasks ("1:T"). All results are presented as averages across 3 seeds which control the task split. Standard deviations are omitted since they represent variance across task splits, not variance across runs of each algorithm.

For each training batch, we allow replay samples from all previous classes, which are concatenated with samples from the current task to make up the training batch. When selecting samples to store in this replay buffer, we follow the strategy detailed in \cite{Hickok2024WatchYS}. When retrieving samples from the buffer for replay, we sample from a uniform distribution across all previously encountered classes. Furthermore, to promote sample diversity, we ensure that no duplicate samples are retrieved within any given stretch of 4000 retrieved samples, again supported by \cite{Hickok2024WatchYS}.

In the class-incremental setting, this replay mechanism involves retrieving samples from all previous downstream tasks. In continual pre-training, this means retrieving from all previous downstream tasks along with samples from the pre-training dataset. We define the replay ratio (RR) as the ratio of replay samples to samples from the current task:
\[
\mathrm{RR} = \frac{N_{\mathrm{replay}}}{N_{\mathrm{task}}},
\]
where $N_{\mathrm{replay}}$ is the number of replayed samples and $N_{\mathrm{task}}$ the number of current-task samples. By default, we use a 1.0 replay ratio. Note that, since the total training batch size is fixed, using a smaller replay ratio means more of each training batch is dedicated to samples from the current task; in this case, total training time is reduced as fewer total samples are used for training (see Figure \ref{fig:rr_graphic} for a visual explanation). For more details (including our loss function), see the appendix.

\begin{figure}[t] % Use figure* instead of figure to span both columns
                   % [t] suggests placement at the top of a page
    \centering     % Center the image within the full page width
    \includegraphics[width=0.9\linewidth]{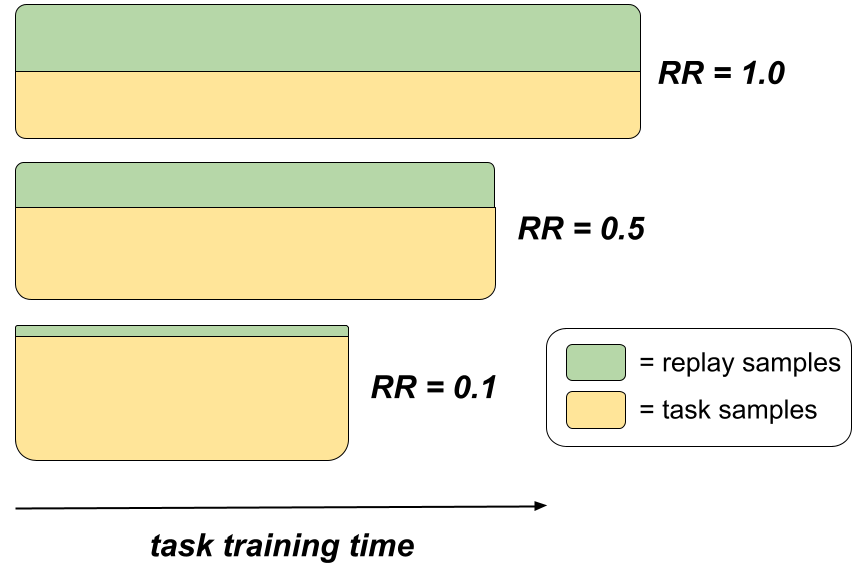} % \linewidth now refers to the full text width
    \caption{Visual representation of the impact of changing the replay ratio (RR). Each block represents the samples used during learning of a single task. Reducing the replay ratio allows us to allocate more of the training batch to task samples, which reduces the total number of samples processed and, therefore, reduces training time.}
    \label{fig:rr_graphic}
\end{figure}

\begin{figure*}[ht] % Use figure* for two-column spanning, [t] suggests top placement
    \centering % Center the content within the full width
    \begin{subfigure}[t]{0.9\linewidth} % Adjust width to slightly less than half
        \includegraphics[width=\linewidth]{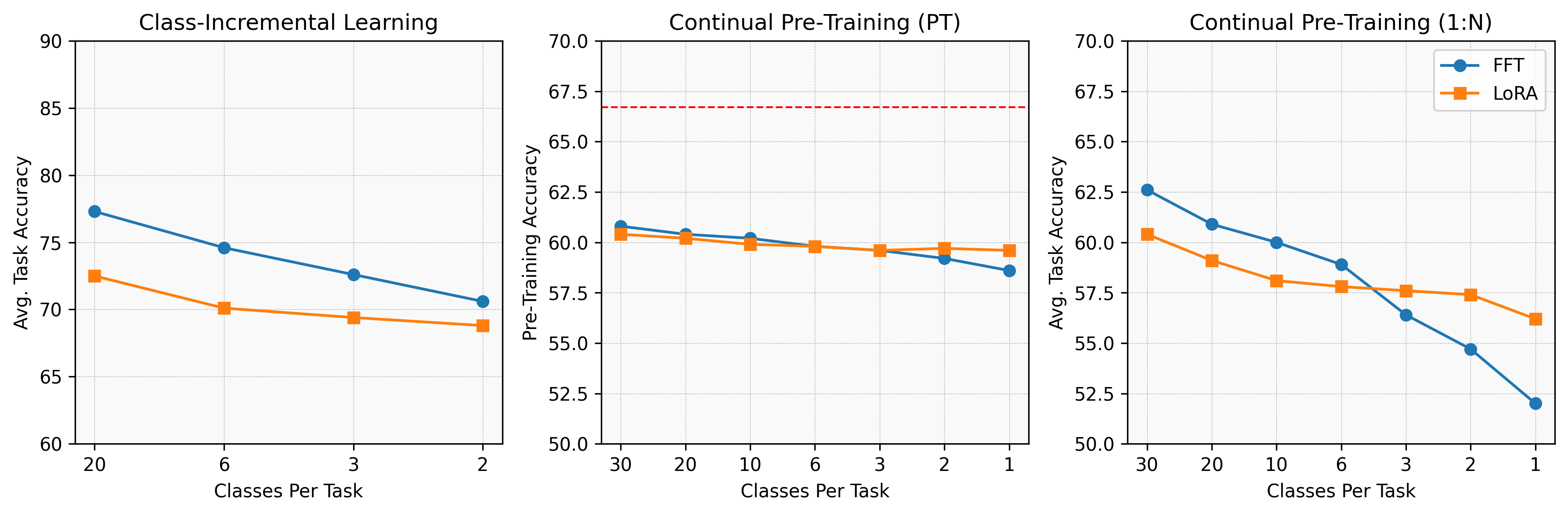}
        % \caption{Description of the first image} % Optional caption for subfigure 1
        \label{fig:peft_task_size_sub} % Label for the subfigure 1
    \end{subfigure}% <--- Keep the % here to avoid unwanted space
    \hfill % Adds flexible space between the subfigures
    \begin{subfigure}[t]{0.9\linewidth} % Adjust width to slightly less than half
        \includegraphics[width=\linewidth]{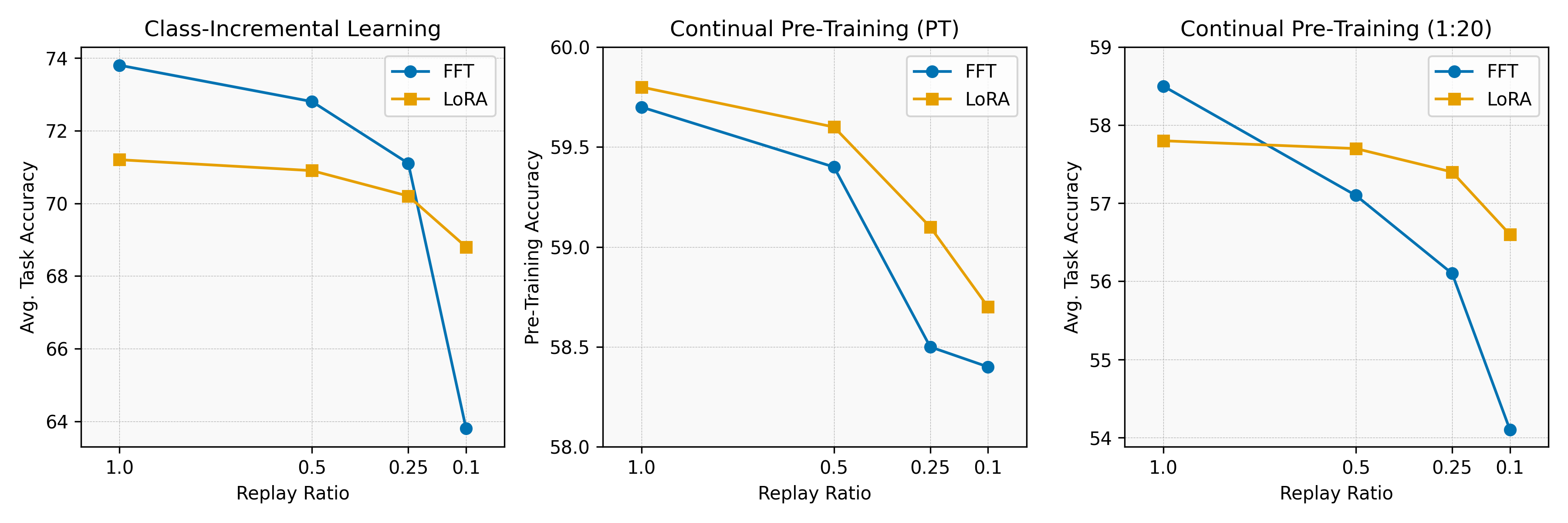}
        % \caption{Description of the second image} % Optional caption for subfigure 2
        \label{fig:peft_rr_sub} % Label for the subfigure 2
    \end{subfigure}
    \caption{LoRA performance vs. FFT performance across task sizes while keeping total downstream classes fixed at 120 (top). LoRA performance vs. FFT performance across replay ratios (bottom). Note that "Continual Pre-Training (PT)" indicates pre-training accuracy at the end of training while "Continual Pre-Training (1:N)" indicates average downstream task accuracy at the end of training.}
    \label{fig:combined_peft} % Label for the combined figure
\end{figure*}
\section{Low Rank Adaptation}
\label{sec:lora}

Parameter-Efficient Fine-Tuning (PEFT) methods have become popular for adapting large pre-trained models to downstream tasks without modifying the bulk of the original model parameters \cite{han2024parameter}. Among these, Low-Rank Adaptation (LoRA)~\cite{hu2021lora} has emerged as a prominent technique. LoRA operates on the principle that the change in model weights during adaptation ($\Delta W$) can be effectively approximated by a low-rank decomposition. Specifically, for a pre-trained weight matrix $W_0 \in \mathbb{R}^{d \times k}$, LoRA introduces two smaller matrices, $A \in \mathbb{R}^{d \times r}$ and $B \in \mathbb{R}^{r \times k}$, where the rank $r \ll \min(d, k)$. The adapted weight matrix $W$ is then represented as $W = W_0 + BA$. During fine-tuning, only the parameters of $A$ and $B$ are trained, significantly reducing the number of trainable parameters compared to updating the full matrix $W_0$.

\subsection{Applying LoRA to Continual Learning}

Unlike typical LoRA usage where adapters might be kept separate for task-switching, we adapt the methodology to allow knowledge accumulation directly into the base model parameters over a sequence of tasks. For each incoming task $t$, we initialize a new LoRA module (matrices $A_t$ and $B_t$) for each attention module and MLP layer of the current model $M_{t-1}$ (we tried several LoRA insertion configurations). We then fine-tune only these LoRA parameters on the data for task $t$. Upon completion of training for task $t$, we merge the learned low-rank update directly into the base model weights: $W_t = W_{t-1} + B_t A_t$. This yields the updated model $M_t$. Subsequently, the LoRA parameters $A_t$ and $B_t$ are discarded, and a fresh set of LoRA parameters ($A_{t+1}, B_{t+1}$) are initialized for the next task, $t+1$.

LoRA offers attractive properties in this context: it is memory-efficient, computationally lightweight, and—due to the representational constraints imposed by the low-rank updates—it provides a form of regularization. In essence, the low-rank adaptation imposes a capacity limit on how much the model can change per task. This means that LoRA tends to \emph{learn less} new information from each task (relative to unconstrained full fine-tuning) and, as a direct result, \emph{forget less} of the knowledge acquired from previous tasks. By restricting $\Delta W$ to a limited subspace, LoRA prevents large weight deviations that could overwrite prior task solutions, thereby serving as a built-in hedge against catastrophic forgetting. However, if a new task’s requirements lie largely outside the subspace spanned by the low-rank adapters, LoRA may underfit the new task due to insufficient adaptation capacity. We thus investigate whether the benefits of this trade-off manifest as improved performance when using LoRA in place of standard full-model fine-tuning (FFT) across various continual learning scenarios, task distributions, and replay conditions.

\subsection{LoRA vs FFT}

In Figure~\ref{fig:combined_peft} (top), we compare LoRA and FFT across different task sizes while keeping the total number of downstream classes fixed at $T \times C = 120$. For example, 30 classes per task implies $T=4$ tasks in sequence, whereas 2 classes per task implies $T=60$ sequential tasks. In the CIL setting, we find that LoRA underperforms FFT by a wide margin for all but the most adversarial scenario of extremely small tasks (even in that extreme case, LoRA is still outshined by FFT). In the CPT setting, by contrast, LoRA begins to outperform FFT once tasks become sufficiently small (around 3 classes per task), and it remains on par with FFT in maintaining pre-training accuracy across all task sizes. Figure~\ref{fig:combined_peft} (bottom) shows another LoRA-vs-FFT comparison, this time varying the replay ratio while fixing the sequence length at 20 tasks (with 6 classes each). In the CPT setting, LoRA clearly outperforms FFT at any replay ratio below 1.0, effectively preventing catastrophic forgetting as the replay ratio approaches zero. In the CIL setting, the greater plasticity of FFT allows it to continue outperforming LoRA down to a replay ratio of about 0.25, although FFT’s performance drops off dramatically once the replay ratio reaches 0.1 (at which point LoRA has a notable advantage).

These results have multiple implications. First, it is clear that smaller tasks inevitably lead to reduced performance; dividing the stream into many tiny tasks means more frequent weight updates and fewer samples per task, which makes learning less stable and increases interference. Second, these results suggest that LoRA is optimal in cases where retention demands are high (e.g in the CPT setting, where over 600 classes cannot be forgotten) and in cases where the model does not need more representational power than what LoRA affords (e.g in the case of small tasks). Third, LoRA is especially effective at preventing catastrophic forgetting when the amount of replay is extremely limited. As the replay ratio decreases, an FFT-based model quickly forgets earlier tasks (indeed, we observe FFT performance collapsing when replay is only 0.1), whereas the LoRA-based model degrades far more gracefully. We will leverage this robustness in Section~\ref{sec:consol} by introducing a consolidation phase to compensate for a reduced replay ratio. Importantly, these results are not simply due to an insufficient LoRA rank ($r$), as we tested each rank in {4, 8, 16, 32, 64, 128}, finding $r=8$ to perform best for task containing between 1 and 6 classes and $r=32$ to perform best for tasks with 10 through 30 classes.

\textbf{How will it scale?:} besides its memory and computational efficiency, LoRA will become more effective as models and datasets scale since the primary drawback of LoRA (insufficient information capacity) will become less of an issue as models have more skills to provide a foundation for learning new tasks \cite{zhang2024when}; practitioners will be able to enjoy the benefits of efficiency and regularization without having to sacrifice performance. On the other hand, in cases where each bout of learning is large and/or complex (which is ideal for continual learning) and where compute resources are abundant, FFT may be a better option given the use of an effective replay algorithm. This is supported by FFT outperforming LoRA for both large tasks in the CPT setting and across all task sizes in the CIL setting, and is further supported by the intuition that the supervision provided by the replay loss (which leverages knowledge distillation over logits from expert model checkpoints) will scale quite graciously: logits will increase in their fidelity/calibration as we scale data and model parameters \cite{Xie2020SelftrainingWN, Busbridge2025DistillationSL}.

\begin{figure*}[t] % Use figure* instead of figure to span both columns
                   % [t] suggests placement at the top of a page
    \centering     % Center the image within the full page width
    \includegraphics[width=0.8\linewidth]{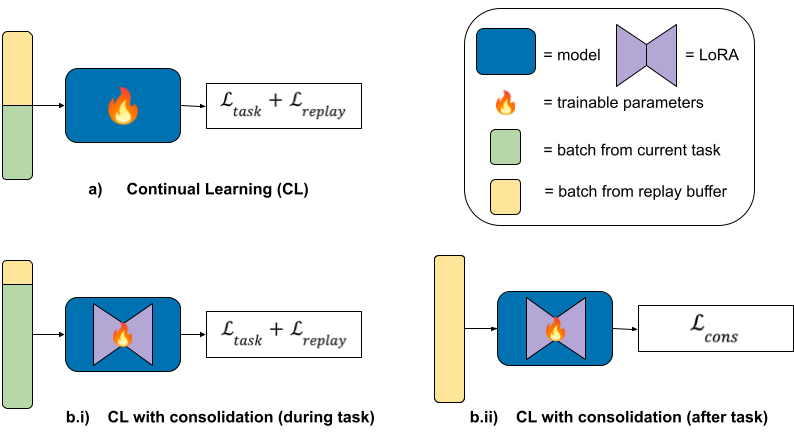} % \linewidth now refers to the full text width
    \caption{Visual representation of continual learning with consolidation. Standard continual learning complements task samples with many replay samples per batch, and does not have a post-task training phase. With consolidation, we reduce the replay ratio during task learning (optionally using LoRA depending on the training distribution) then execute a post-task training phase on a distribution optimized for reconciliation of old and new knowledge. This allows a vastly reduced number of total gradient steps across training.}
    \label{fig:consol_graphic}
\end{figure*}

\section{Consolidation}
\label{sec:consol}

A key challenge in continual learning is mitigating catastrophic forgetting without incurring excessive computational costs. While replay is a widely adopted and effective strategy, the number of replayed samples can become a bottleneck, especially in long learning sequences or resource-constrained environments. To address this, we introduce a novel consolidation strategy designed to drastically improve the sample efficiency of replay. The core idea is to drastically reduce the replay ratio during the learning phase of a new task and then dedicate a separate, subsequent phase to consolidate all existing knowledge, focusing on reinstating information that may have degraded while learning the most recent task.

\subsection{Method}

Our approach involves two main stages. First, when learning a new task, the model is trained with a replay ratio (RR, defined in Section \ref{sec:background}) that is substantially lower than the typical 1:1 baseline (where the number of replayed samples equals the number of current task samples). This initial reduction in replay during task acquisition lessens the immediate computational burden and leads to a manageable decrease in post-task performance (see Figure \ref{fig:combined_peft}), especially when regularizing with LoRA. Second, after the model has been trained on the current task, it enters a consolidation phase. During this phase, the model is exclusively trained on replayed samples that drawn from a distribution optimized to reinstate any knowledge that has degraded while learning the task. While this distribution could directly sample from classes that have seen drops in accuracy (or some kind of feature/prototype drift) during training, our work serves as a proof-of-concept; we use a balanced distribution across all previous classes during the consolidation period. Besides this distribution, future work can also develop loss functions that are specialized for the consolidation period; here, we use the same replay loss function that is used during task learning (cross entropy + logit distillation).

To quantify and control the amount of replay utilized during the consolidation phase, we define the consolidation step rate (CSR). The CSR is a value between $0$ and $1$ that determines the proportion of "saved" replay samples—those that would have been used if a standard $1:1$ replay ratio was maintained during task learning—that are instead allocated to training steps during the consolidation phase. For example, if reducing the replay ratio from $1.0$ to $0.25$ during task learning saves $N_{\text{saved}}$ potential replay samples compared to the $1:1$ baseline, a CSR of $0.5$ would mean that $0.5 \times N_{\text{saved}}$ samples are used for training during the consolidation period.

To evaluate the overall sample efficiency of our method, we define the total replay percentage (TRP). The TRP measures the total percentage of replay samples used across all tasks (including both the task-learning phase and the consolidation phase) relative to the number of replay samples that would have been used with a standard $1:1$ replay ratio throughout the entire continual learning sequence. For instance, employing a replay ratio of $0.25$ during task learning and a CSR of $1.0$ (meaning all saved replay samples are used for consolidation) would result in a TRP of $100\%$, as the total number of replay samples used ($N_{\text{replay\_task}} + N_{\text{replay\_consolidation}}$) matches that of a standard $1:1$ approach, just redistributed. By varying the RR and CSR, our goal is to achieve a TRP significantly below $100\%$ while maintaining comparable or even improved performance. The TRP is calculated as:
$$\text{TRP} = \frac{\sum_{i=1}^{T} (N_{\text{replay\_task},i} + N_{\text{replay\_consolidation},i})}{\sum_{i=1}^{T} N_{\text{replay\_baseline\_1:1},i}} \times 100\%$$

\newcolumntype{S}[1]{>{\raggedright\arraybackslash}p{#1}}

\begin{figure*}[t]
  \centering
  %-----------------------------------------------------------%
  %--------------------------- TABLE --------------------------%
  %-----------------------------------------------------------%
  \begin{subfigure}[t]{0.55\textwidth}
    \centering
    % tighten row spacing in the table if desired
    \renewcommand{\arraystretch}{1.15}
    % --------- table itself --------- %
    \begin{tabularx}{\linewidth}{|S{0.121\textwidth}|S{0.121\textwidth}|S{0.121\textwidth}|>{\columncolor{gray!10}}S{0.121\textwidth}|S{0.121\textwidth}|S{0.121\textwidth}|}
      \hline
      & & & \multicolumn{1}{|>{\columncolor{gray!10}}c|}{\textbf{CIL}} & \multicolumn{2}{|c|}{\textbf{CPT}} \\
      \hline
      \textbf{RR} & \textbf{CSR} & \textbf{TRP} & \textbf{1:20} & \textbf{PT} & \textbf{1:20} \\
      \hline
      0.0  & 0.45 & 45\%  &  --  & 59.7 & 58.1 \\
      0.1  & 0.39 & 45\%  & 72.8 & 59.7 & 58.7 \\
      0.25 & 0.27 & 45\%  & 74.2 & 59.7 & 58.6 \\
      0.45 & 0.00 & 45\%  & 72.6 & 59.5 & 57.9 \\
      \hline
      0.0  & 0.55 & 55\%  &  --  & 59.8 & 58.2 \\
      0.1  & 0.50 & 55\%  & 73.3 & 59.7 & 59.1 \\
      0.25 & 0.40 & 55\%  & 74.6 & 59.7 & 58.7 \\
      0.5  & 0.10 & 55\%  & 73.6 & 59.8 & 58.5 \\
      0.55 & 0.00 & 55\%  & 72.8 & 59.5 & 57.8 \\
      \hline
      0.0  & 1.00 & 100\% &  --  & 59.9 & 59.1 \\
      0.1  & 1.00 & 100\% & 75.0 & 60.0 & 59.5 \\
      0.25 & 1.00 & 100\% & 76.1 & 60.1 & 59.7 \\
      0.5  & 1.00 & 100\% & 75.5 & 60.0 & 59.6 \\
      1.0  & 0.00 & 100\% & 73.8 & 59.7 & 58.5 \\
      \hline
    \end{tabularx}
    % \caption*{\textbf{Hyper-parameter sweep}}  % caption *inside* subfigure (optional)
    \label{tab:results}
  \end{subfigure}
  \hfill
  %-----------------------------------------------------------%
  %----------------------- VIOLIN PLOTS -----------------------%
  %-----------------------------------------------------------%
  \begin{subfigure}[t]{0.40\textwidth}
    \centering
    \includegraphics[width=\linewidth]{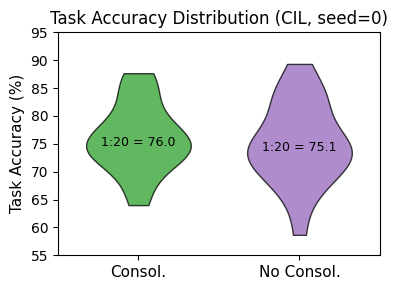}\par\vspace{4pt}
    \includegraphics[width=\linewidth]{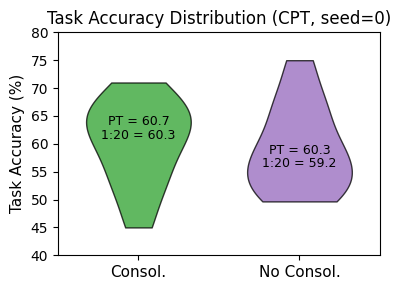}
    % \caption*{\textbf{Per-task accuracy distributions}}
  \end{subfigure}

  %----------------------- GLOBAL CAPTION ---------------------%
  \caption{Performance of standard replay (\textit{no} consolidation) versus consolidation (\textbf{CIL/CPT}).  
           Left: mean final accuracy across tasks for each replay-ratio (\textbf{RR}) and consolidation-sample-ratio (\textbf{CSR}) setting, with \textbf{TRP} reporting the total proportion of replayed samples. Note that a CSR of 0.0 indicates standard replay, which acts as a baseline.  
           Right: violin plots of per-task accuracies at the end of training.}
  \label{fig:consol_violin}
\end{figure*}

In our experiments, we keep the number of tasks fixed at 20 and the number of classes per task fixed at 6; note that the choice of task size dictates whether LoRA is necessary for preventing forgetting during the task learning and/or consolidation phase. In cases where LoRA is necessary (which is true in the continual pre-training (CPT) setting but not the CIL setting), we treat the consolidation phase as another task and follow the same "reinitialize-merge" procedure outlined in Section \ref{sec:lora}; this means LoRA is applied separately during both the task learning and consolidation phases, as illustrated in Figure \ref{fig:consol_graphic}. 

\subsection{Consolidation vs Standard Replay}

Table \ref{fig:consol_violin} shows performances across consolidation step rates, total replay percentages, and continual learning settings. It can be seen that using an equal amount of replay samples for consolidation leads to strong performance improvements over the baseline (which simply matches the RR to the TRP) and, strikingly, that consolidation allows a model trained with up to 55\% less total replay samples to match the 1.0 RR baseline, which is the standard approach found in the literature.

Figure \ref{fig:consol_violin} shows the distributions of task accuracies (i.e the accuracy of each task) for consolidation with a 100\% TRP versus no consolidation and a 1:1 replay ratio. In the class-incremental learning setting, consolidation has the effects of reducing the frequency of outliers and raising the average accuracy. In the continual pre-training setting, consolidation raises PT and task accuracy but does not reduce the frequency of outliers; this underscores the importance of performing more targeted replay during the consolidation phase, which we leave to future work.

While our approach is imperfect, these results support the notion that consolidation can serve as an effective framework for enhancing the sample efficiency and overall effectiveness of replay.

\textbf{How will it scale?:} we hypothesize that the procedure we have outlined (and its derivatives) will prove to be a highly scalable approach to continual learning, again due to the fact that the quality of the logits used for knowledge distillation will improve with scale \cite{Xie2020SelftrainingWN, Busbridge2025DistillationSL}. We imagine a scenario where a near-zero replay ratio is sufficient for regularizing task learning (thanks to highly calibrated logits and minimal weight perturbations necessary for learning new tasks) and where consolidation is done in a highly targeted manner (and with early stopping). We also note that the consolidation phase can be executed during off-peak datacenter hours and, as a final practical suggestion, we recommend delaying the consolidation phase for as many task learning sessions as possible, as this increases the size of each task and therefore will lead to better results (we verified this empirically but do not have polished results; see Section \ref{sec:lora} for task size effects).

\begin{figure*}[ht] % Use figure* instead of figure to span both columns
                   % [t] suggests placement at the top of a page
    \centering     % Center the image within the full page width
    \includegraphics[width=0.85\linewidth]{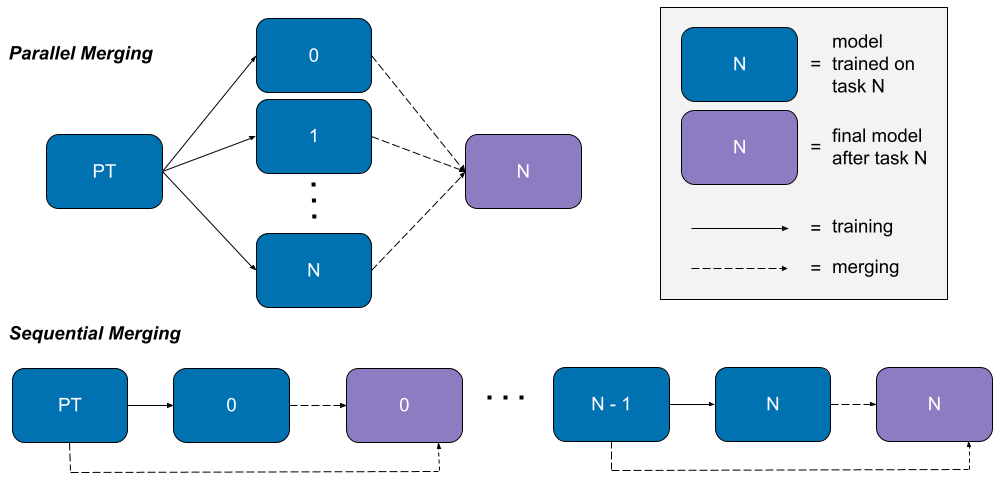} % \linewidth now refers to the full text width
    \caption{Visual representation of the differences between parallel and sequential merging. Parallel merging trains $N$ independent, task-specific models based on the same pre-trained backbone, then merges their weights to produce a model that can perform tasks $1-N$. Sequential merging trains on each task sequentially, merging the pre- and post-task model checkpoint after training on each task.}
    \label{fig:merge_graphic}
\end{figure*}

\section{Model Merging}
\label{sec:merge}

% - Define EMA -> parallel (including replay) -> sequential
% - Experimental setup
% - Hyperparameter sweep

% - Discuss results

% - Why should merging scale well?

In the multi-task fine-tuning literature, model merging (originally known as task arithmetic) has emerged as a surprisingly effective post‑hoc strategy for composing multiple task‑specialized checkpoints into a single set of weights that recovers most of the per‑task accuracy of individual fine‑tunes while avoiding the compute and memory overhead of joint multitask training \cite{yang2024model}. By treating the parameter updates learned for each task as vectors in weight space and averaging—or otherwise linearly combining—those vectors, practitioners can achieve aggregate performance that is typically within one‑to‑two percentage points of the best single‑task models, all with a single forward pass at inference time. Moreover, model merging typically leaves pre-trained knowledge intact, especially when using advanced techniques like DARE-TIES-Merging \cite{Yu2024LanguageMA, Yadav2023TIESMergingRI}. 

Its ability to cheaply integrate specialized models into a unified model while avoiding interference with pre-trained knowledge makes model merging a clear candidate for application to the continual learning setting where, in contrast to the usual multi-task fine-tuning setup, there is a large (indefinite) number of tasks that are encountered sequentially. In this section, we study three algorithms that fit under the "model merging" umbrella: parallel merging (task arithmetic) \cite{ilharco2022editing}, exponential moving average (EMA) \cite{soutifcormerais2023improving, gao2023lae}, and a sequential merging strategy that is meant to account for the extensive and sequential nature of the task streams encountered during continual learning. Throughout our experiments, we continue using replay, which is further distinct from the multi-task fine-tuning literature.

\subsection{Methods}

In this section, we formalize each merging algorithm. See Figure \ref{fig:merge_graphic} for a visual comparison of parallel and sequential merging.

\textbf{Parallel Merging:} parallel merging, sometimes called \textit{task arithmetic} treats each task’s fine‑tuning as a \emph{vector} in parameter space \cite{ilharco2022editing}. Starting from the pre‑trained weights $\theta_{0}$, we fine‑tune an independent copy $\theta_{t}^{\star}$ on every task $t!\in!{1,\dots,T}$ and define the task vector $\tau_{t}=\theta_{t}^{\star}-\theta_{0}$. After all tasks are learned, we synthesize a single multi‑task model by adding a weighted sum of these task vectors back to the base model

$$
\theta_{\text{PM}}
\;=\;
\theta_{0}
\;+\;
\sum_{t=1}^{T}\alpha_{t}\,\tau_{t},
$$

where the coefficients $\alpha_{t}$ are commonly uniform ($\alpha_{t}=1/T$) but can be tuned to balance tasks (we tune a single $\alpha$ which is used for all $\tau_{t}$). Empirically, this procedure delivers strong multi‑task performance and can be enhanced with scaling, sparsification, or trust‑region projections to mitigate interference among task vectors; here, we use DARE-TIES \cite{Yu2024LanguageMA, Yadav2023TIESMergingRI} merging, tuning the dropout probability of DARE. This strategy is a standard in multi-task fine-tuning, especially when the goal is to maintain the capabilities of the pre-trained model while incorporating new task-specific knowledge. In our case, a linear probe is trained at the end of the training phase to allow us to the assess the ability of the learned representations to support image classification.

When combining parallel merging with replay, we allow replay from the pre-training dataset and/or from previous tasks. In the case of replay purely from the pre-training dataset, the pre-trained classifier is not re-initialized; with replay from the pre-training dataset and from previous tasks, we add neurons for each sequentially encountered task; in the case of no replay, we initialize an independent classifier for each task.

\begin{table}[t]
\centering
\small
\begin{tabularx}{0.4\textwidth}{|>{\raggedright}m{2.5cm}|X|X|}
\hline
% & \multicolumn{2}{|c|}{\textbf{CPT}} \\
% \hline
% & \multicolumn{2}{|c|}{\textbf{20x6}} \\
% \hline
& \textbf{PT} & \textbf{1:20} \\
\hline
Parallel (NR) & 58.9 & 58.2 \\
\hline
Parallel (PT) & 59.4 & 58.8 \\
\hline
Parallel (PT + 1:20) & 58.6 & 58.9 \\
% + TIES & 72.0 \tiny{±1.4} & 76.9 \tiny{±3.4} & 87.6 \tiny{±3.4} \\
\hline
\end{tabularx}
\caption{Results for each type of replay possible with parallel merging. "NR" indicates No Replay, "PT" indicates replay from the Pre-Training dataset, and "PT + 1:20" indicates replay from the Pre-Training dataset and from tasks 1 through 20 (from previously seen tasks).}
\label{table:parallel_replay}
\end{table}

\begin{figure*}[t] % Use figure* instead of figure to span both columns
                   % [t] suggests placement at the top of a page
    \centering     % Center the image within the full page width
    \includegraphics[width=1.0\linewidth]{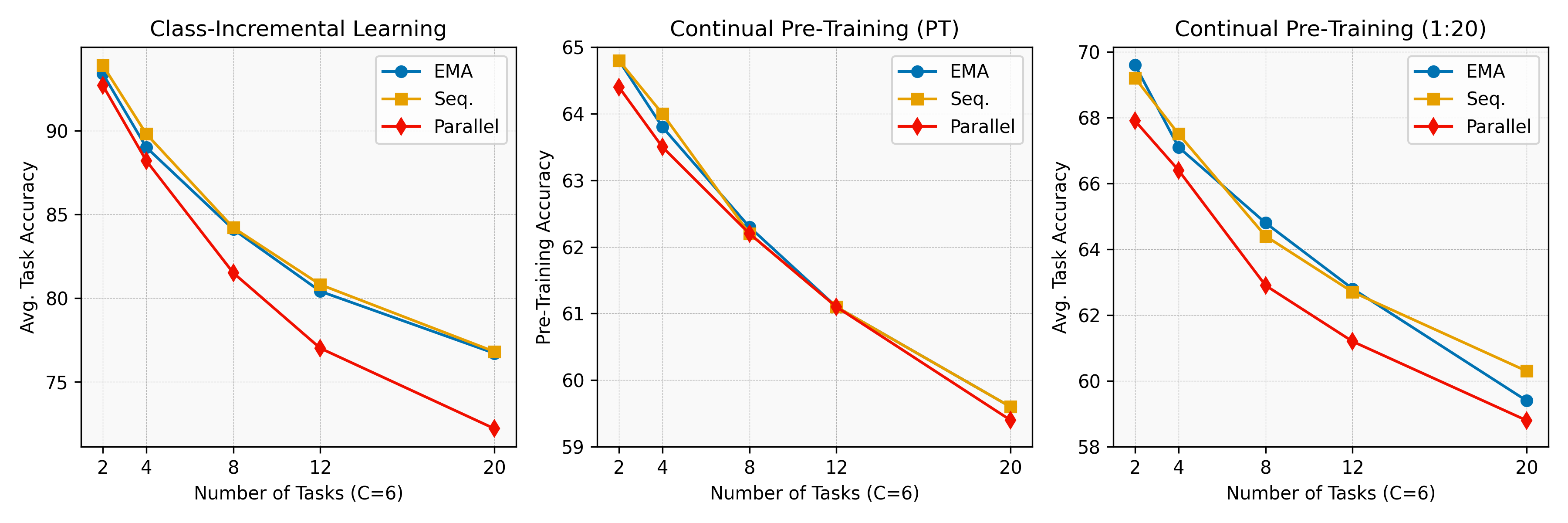} % \linewidth now refers to the full text width
    \caption{Performances of parallel merging, EMA, and sequential merging while varying the number of downstream tasks, keeping the number of classes per task fixed at 6. Note that "Continual Pre-Training (PT)" indicates pre-training accuracy at the end of training while "Continual Pre-Training (1:N)" indicates average downstream task accuracy at the end of training.}
    \label{fig:merge_results}
\end{figure*}

\textbf{Exponential Moving Average (EMA):} a mainstay of modern optimization, Exponential Moving Average (EMA) offers an effective, low-overhead strategy for ensembling model parameters throughout the training trajectory. Its utility in continual learning is particularly pronounced, where EMA serves as an implicit regularizer, encouraging smoother transitions in the parameter landscape as the model adapts to sequential data \cite{soutifcormerais2023improving, gao2023lae}.

Given the model parameters $\theta_k$ after the $k$-th gradient step, the EMA maintains a "shadow" parameter set $\theta_{\text{EMA}}$, which is iteratively refined:
$$
\theta_{\text{EMA},k} \;=\; \lambda\:\theta_{\text{EMA},k-1} \;+\; (1-\lambda)\:\theta_{k}
$$

with $\theta_{\text{EMA},0}$ initialized to $\theta_{0}$. The decay coefficient $\lambda \in [0,1)$ controls the memory of the averaging process, determining the influence of historical parameters. EMA's online update mechanism is computationally efficient, demanding only a single duplicate of the model's parameters and operating seamlessly across task boundaries-a significant advantage for many continual learning settings. At the end of each task, the standard parameters $\theta_k$ are superseded by $\theta_{\text{EMA}}$; EMA continues from these parameters upon entering the next task.

\textbf{Sequential Merging:} to directly address the lengthy and sequential nature of task streams in continual learning, we propose sequential merging. Upon the completion of training for task $t$, the model weights $\theta_{t}^{\star}$, which have been adapted for this specific task, are merged with the accumulated knowledge represented by the pre-task weights $\theta_{t-1}$:
$$
\theta_{t} \;=\; (1-\alpha)\,\theta_{t-1} \;+\; \alpha\,\theta_{t}^{\star}
$$
where the hyperparameter $\alpha \in (0,1]$ governs the integration process. Sequential merging is highly efficient, especially since, unlike EMA, we can tune the merging hyperparameter ($\alpha$) without having to either re-train the model or store a checkpoint after each gradient step. This is important since $\alpha$ is likely highly dependent on the model's current capabilities relative to the given downstream task, as well as application-specific performance tradeoffs.

\subsection{Parallel Merging with Replay}

Parallel merging (and, to the best of our knowledge, other forms of merging) has not been combined with replay up to this point, and the ability to sidestep the need for replay (and continual learning methods in general) is seen as one of its key benefits; however, there is the possibility that replay could push task-specialized models ($\theta_{t}^{\star}$) into states that are more favorable for merging. We take it as a given that replay combines well with sequential merging and EMA, which we confirmed with a replay ratio sweep (see Table \ref{table:seq_replay}).

Table \ref{table:parallel_replay} shows a comparison of the accuracies of each type of replay-enabled parallel merging. Notably, replaying samples from the pre-training dataset leads to a noticeable boost in pre-training accuracy and task accuracy, suggesting that explicitly optimizing each task-specialized checkpoint for maintenance of pre-trained knowledge can be a worthwhile investment. Conversely, replaying from both pre-trained classes and task classes does not lead to improved performance, which is unsurprising given the known need for task vectors to operate in independent, localized subspaces of parameter space \cite{OrtizJimenez2023TaskAI}.

\subsection{Comparing Methods}

We now turn to the question of whether sequential merging is able to outperform parallel merging in continual learning settings. We hypothesize that parallel merging will inevitably break down in the presence of many tasks, as the probability of each task vector occupying independent, localized subspaces is reduced with each additional task (we rely on DARE-TIES merging \cite{Yu2024LanguageMA, Yadav2023TIESMergingRI} to counteract this effect). Another object of interest is the relative performances of EMA and sequential merging, as EMA serves as a strong baseline \cite{soutifcormerais2023improving, gao2023lae} that belongs to the broader family of "sequential merging" algorithms (since training is only based on the pre-trained checkpoint at the beginning of the first downstream task).

Figure \ref{fig:merge_results} shows a comparison of the accuracies of parallel merging, sequential merging, and EMA while varying the number of tasks and the continual learning setting; we keep the number of classes per task fixed at 6 meaning that, as more tasks are added, the total number of classes increases. In the class-incremental learning setting, parallel merging increasingly falls behind sequential merging and EMA (which perform similarly) as more tasks are added; at 2 tasks, parallel merging is competitive; at 20 tasks, there is a ~5\% difference between sequential and parallel merging. In the continual pre-training setting, we see a similar trend, although unlike in the CIL setting the gap between parallel and sequential merging does not dramatically increase as more tasks are added.

These results suggest that sequential merging is better suited to the continual learning setting than parallel merging and can match the strong EMA baseline. Given real-world demands for a cheap merging algorithm that scales to many sequential tasks, we expect variants of sequential merging (combined with replay) to become reliable tools for future systems.

\textbf{How will it scale?:} on top of the outright efficiency of model merging, recent work \cite{Yadav2024WhatMF} suggests that parallel model merging increases in effectiveness as models and datasets scale. In particular, Yadav et al. show that, for large‑capacity models and richer pre‑training corpora, the vectors learned by each task become both lower‑norm and more orthogonal, so their simple linear combination preserves per‑task optima with vanishing interference. We believe these properties will also hold true in the case of sequential merging, with the added benefits of new learning being able to dynamically configure the parametric subspaces occupied by the current and previous tasks all while completely avoiding the basic model capacity constraints that parallel merging runs into.
\begin{table}[t]
\centering
\small
\begin{tabularx}{0.482\textwidth}{|S{0.14\textwidth}|S{0.054\textwidth}|>{\columncolor{gray!10}}S{0.054\textwidth}|S{0.054\textwidth}|S{0.054\textwidth}|}
\hline
& & \multicolumn{1}{|>{\columncolor{gray!10}}c|}{\textbf{CIL}} & \multicolumn{2}{|c|}{\textbf{CPT}} \\
\hline
& \textbf{TRP} & \textbf{1:20} & \textbf{PT} & \textbf{1:20} \\
\hline
No merge/consol. & 35\% & 72.0 & 59.3 & 57.9 \\
Consol. & 35\% & 73.1 & 59.5 & 58.3 \\
Parallel & 35\% & - & 59.4 & 58.7 \\
Parallel + Consol. & 35\% & 72.5 & 59.5 & 58.8 \\
Seq. & 35\% & 73.5 & 59.4 & 58.4 \\
Seq. + Consol. & 35\% & 73.8 & 59.8 & 58.8 \\
\hline
No merge/consol. & 100\% & 73.8 & 59.7 & 58.5 \\
Consol. & 100\% & 76.1 & 60.1 & 59.7 \\
Parallel & 100\% & - & 59.4 & 58.8 \\
Parallel + Consol. & 100\% & 73.7 & 59.6 & 58.8 \\
Seq. & 100\% & 76.0 & 59.9 & 59.8 \\
Seq. + Consol. & 100\% & 76.5 & 60.4 & 60.0 \\
\hline
\end{tabularx}
\caption{Performance comparison of combined strategies, including "Seq. + Consol." which combines our two main methods, sequential merging and consolidation. LoRA is used situationally, only being used here in the case of consolidation with no merging.}
\label{table:ablate}
\end{table}

\section{Combining Strategies}
\label{sec:combining}

% - Describe how they're combined (as necessary)

% - Discuss results

In this section, we wrap up our study by determining whether each strategy listed so far can be combined to outperform standalone variants. We also seek to determine whether combining strategies can lead to furthermore sample-efficient replay.

To combine parallel merging and consolidation, we use a near-zero replay ratio while fine-tuning each independent model (which we found to perform comparable to a 1.0 replay ratio) then allocate saved replays to a consolidation phase that is executed after the linear probe is trained to convergence. Note that the difference between "Parallel" and "Parallel + Consol." ends up being quite small due to the fact that the linear probe is already trained on a balanced distribution. Combining sequential merging and consolidation is straightforward; after each task, we merge the pre- and post-task model then execute the consolidation phase. We keep the number of tasks fixed at 20 and the number of classes per task fixed at 6. For each algorithmic configuration, we tune all available hyperparameters, which includes loss coefficients, merging $\alpha$, consolidation step rates, replay ratios, and LoRA rank (if LoRA is being used).

Table \ref{table:ablate} shows a comparison of each configuration, complete with total replay percentages (TRP). Combining sequential merging and consolidation proves to be the optimal strategy, with this combination reaching the highest accuracies across the CIL and CPT settings. We also see that using this combination allows us to reach the same performance as our baseline ("No merge/consol.") while using 65\% less replay samples, which is less than standalone consolidation requires for the same performance (see Table \ref{table:consol}). These results confirm our suspicion that the strategies we have developed can be combined seamlessly, making the combination of consolidation, sequential merging, and (informed) LoRA ready for scaling.
\section{Conclusion}
\label{sec:conclusion}

In summary, we have presented scalable strategies for replay-based continual learning and demonstrated their effectiveness across long task sequences and true continual learning settings. Our study focused on three complementary techniques: low-rank adaptation (LoRA) for efficient and regularized fine-tuning, phasic replay for enhanced sample efficiency, and model merging for further regularization. Empirically, we found that LoRA adapters can significantly boost performance in under-regularized regimes (such as when tasks are highly disjoint or when using very low replay ratios), although standard full-model fine-tuning remains superior given a high enough replay ratio. We showed that a simple post-task consolidation of replay data allows the model to rebalance its knowledge and reduces the total number of replay samples required by up to 55\% to reach a given performance level, a substantial improvement in sample efficiency. Additionally, by merging the model’s weights before and after each task, we achieved performance comparable to maintaining an online exponential moving average, indicating that merging is an effective mechanism for retaining prior task accuracy without continuous overhead. Despite these encouraging results, our work has two key limitations that suggest avenues for future research. First, we limited our experiments to image classification; this is seen as a highly adversarial setting given a lack of smooth boundaries between classes and tasks, which would be present if we used CLIP, for example. Nonetheless, the simplicity and generality of our methods leads us to hypothesize that they will seamlessly transfer to multimodal models. Second, the scalability to extremely large task sequences or model sizes is still unproven. While our methods performed well for on the order of tens of tasks, real-world continual learning deployments might involve hundreds of tasks and/or the continual adaptation of very large vision models (with billions of parameters). While our methods were chosen with scalability in mind, their performance characteristics at that scale remain an open question.
{
    \small
    \bibliographystyle{ieeenat_fullname}
    \bibliography{main}
}

\clearpage
\setcounter{page}{1}
\maketitlesupplementary

\section{Training Details}
\label{sec:setup}

For samples from the current task, we use a standard cross-entropy loss. For replay samples, we use a variant of the DER++ \cite{Buzzega2020DarkEF} objective that replaces raw logit distillation with traditional knowledge distillation (KD) \cite{Hinton2015DistillingTK} and adds logit standardization \cite{Sun2024LogitSI} for normalization. For each training batch, we scale the replay loss then add it to the current task loss. Logits from past classes are not updated throughout training. We also used loss decoupling \cite{Liang2023LossDF} for extra regularization; this technique decomposes the current-task loss into a component that only considers classes from the current task and a component that considers all classes in the classifier (similar to SS-IL \cite{Ahn2021SSILSS}).

We use a cosine learning rate schedule with a linear warmup of 3 epochs, training for 8 epochs total. We sweep the learning rate at each turn, and take the best-performing post-epoch checkpoint at the end of each task (based on validation accuracy).

Beginning in Section \ref{sec:consol}, we switch from uniform replay across all previously encountered \textit{classes} (as described in Section \ref{sec:background}) to uniform replay across all previously encountered \textit{tasks}, allowing us to introduce a hyperparameter that scales the probability of retrieving samples from the original pre-training dataset versus samples from the subsequent $T$ downstream tasks during replay. During both task learning and consolidation, this allows us to balance the influence of samples from the pre-training dataset and from downstream tasks, giving us a clear indication of whether consolidation actually outperforms non-consolidation baselines in terms of both $PT$ and $1:20$ accuracies.

\section{Extra Results}

\begin{table}[h!]
\centering
\small
\begin{tabularx}{0.85\textwidth}{|>{\raggedright}m{0.75cm}|>{\columncolor{gray!10}}X|X|X|X|>{\columncolor{gray!10}}X|>{\columncolor{gray!10}}X|}
\hline
& \multicolumn{2}{|c|}{\textbf{CIL}} & \multicolumn{4}{|c|}{\textbf{CPT}} \\
\hline
& \multicolumn{1}{|c|}{\textbf{N/A}} & \multicolumn{1}{|c|}{\textbf{Seq.}} & \multicolumn{2}{|c|}{\textbf{N/A}} & \multicolumn{2}{|c|}{\textbf{Seq.}} \\
\hline
\textbf{R} & \textbf{1:20} & \textbf{1:20} & \textbf{PT} & \textbf{1:20} & \textbf{PT} & \textbf{1:20} \\
\hline
1.0 & 73.8 & 76.0 & 59.7 & 58.5 & 59.9 & 59.8 \\
\hline
0.5 & 72.8 & 74.7 & 59.6 & 57.7 & 59.7 & 59.0 \\
\hline
0.25 & 71.1 & 72.7 & 59.1 & 57.4 & 59.5 & 58.6 \\
\hline
0.1 & 63.8 & 66.1 & 58.7 & 56.6 & 58.9 & 58.4 \\
% \hline
% 0.05 & - & - & 58.4 \tiny{0.4} & 58.0 \tiny{±1.6} & 58.4 \tiny{±0.4} & 58.7 \tiny{±1.4} & 59.2 \tiny{±0.5} & 58.6 \tiny{±1.9} \\
% \hline
% 0.0 & 11.7 \tiny{±1.7} & 0.0 \tiny{±1.3} & 72.2 \tiny{±1.6} & 21.1 \tiny{±5.0} & 20.3 \tiny{±4.6} & 0.0 \tiny{±0.5} & 0.0 \tiny{±2.2} & 58.9 \tiny{±0.5} & 58.2 \tiny{±1.6} \\
\hline
\end{tabularx}
\caption{Reducing the replay ratio for sequential merging and a no-merging, no-LoRA baseline. Sequential merging combines well with replay.}
\label{table:seq_replay}
\end{table}

\section{Relation to Reinforcement Learning}

Reinforcement learning (RL) is, fundamentally, continual learning. Off-policy RL agents store past interactions in a replay buffer and sample mini-batches of these experiences to break the temporal correlations in continuously generated data, reducing nonstationarity and stabilizing training. This continual rehearsal of diverse past experiences also prevents the agent from forgetting early data and improves its overall sample efficiency. Experience replay is thus a critical component of modern off-policy methods such as DQN, SAC, and TD3, enabling multiple learning updates from each interaction with the environment. A key hyperparameter in these algorithms is the \emph{replay ratio}, defined as the number of gradient updates performed per environment step. High replay ratios (e.g., 4 or more updates per step) are now common in top-performing agents, greatly increasing how frequently past data is re-used for learning. Indeed, recent studies have shown that substantially raising the replay ratio (even an order of magnitude beyond traditional settings) can dramatically improve sample efficiency without degrading performance. In parallel, many works have explored enhancements to the replay mechanism itself, including prioritized experience replay strategies (sampling important transitions more frequently) and adaptive buffer management techniques (e.g., dynamic or limited-size replay buffers) to refresh the stored experience dataset. We mention RL here since our techniques can be directly translated to the RL setting (especially the so-called "continual RL" setting).

\end{document}